%% file: main.tex
\documentclass[10pt,twocolumn,letterpaper]{article}

 \usepackage{cvpr}              %
\usepackage[accsupp]{axessibility} %
\input{preamble}

\definecolor{cvprblue}{rgb}{0.21,0.49,0.74}
\usepackage[pagebackref,breaklinks,colorlinks,citecolor=cvprblue]{hyperref}

\def\paperID{6} %
\def\confName{CVPR}
\def\confYear{2024}

\title{ECLAIR: A High-Fidelity Aerial LiDAR Dataset for Semantic Segmentation}

\author{
Iaroslav Melekhov$^*\,^{1,2}$\quad Anand Umashankar$^*\,^{1}$\quad Hyeong-Jin Kim$^{1}$\quad Vladislav Serkov$^{1}$\quad Dusty Argyle$^{1}$\vspace{8pt}\\
$^{1}$Sharper Shape (\url{https://sharpershape.com/)}\qquad $^{2}$Aalto University
}
\begin{document}

\twocolumn[{%
\renewcommand\twocolumn[1][]{#1}%
\maketitle
    \begin{center}
            \vspace{-0.7cm}
            \captionsetup{type=figure}
            \includegraphics[width=\textwidth]{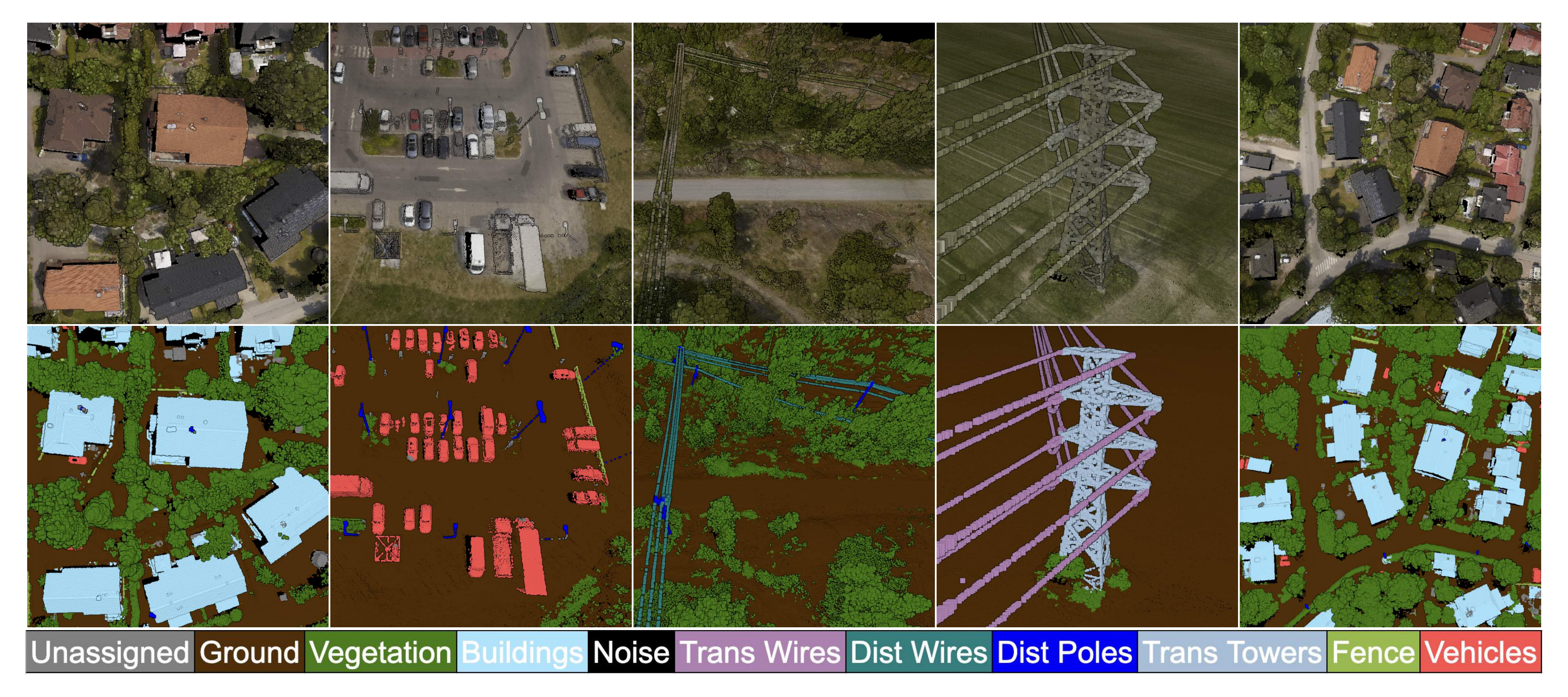}
            \vspace{-0.7cm}
            \captionof{figure}{\textbf{Overview of the proposed ECLAIR dataset.} We introduce ECLAIR, a new outdoor large-scale aerial LiDAR dataset. It covers a total area of more than 10 square kilometers encompassing 11 semantic classes. The long-tail accurate annotations enable fine-grained semantic understanding. Different semantic classes are labeled by different colors.}
    \end{center}
}]

\def\thefootnote{*}\footnotetext{Indicates equal contribution. Correspondence by email: firstname.lastname@sharpershape.com}
\input{sec/0_abstract}    
\input{sec/1_intro}
\input{sec/2_related_work}

\input{sec/3_dataset_method}

\input{sec/4_experiments}

\input{sec/5_conclusion}
{
    \small
    \bibliographystyle{ieeenat_fullname}
    \bibliography{main}
}

\end{document}

%% file: preamble.tex
 \usepackage[dvipsnames]{xcolor}
\usepackage{graphicx}
\usepackage{booktabs}
\usepackage{multirow}
\usepackage{colortbl}
\usepackage{tabularx}
\usepackage{wrapfig}
\usepackage{amsmath}
\usepackage{amssymb}
\usepackage{pifont}
\usepackage{caption}
\usepackage{subcaption}
\usepackage{amsfonts}
\usepackage{rotating} 
\usepackage{makecell}

\newcommand{\cmark}{\ding{51}}%
\newcommand{\xmark}{\ding{55}}%
\newcommand{\greencmark}{{\color{Green}\cmark}}
\newcommand{\redxmark}{{\color{red}\xmark}}

\newcommand{\boldparagraph}[1]{\noindent{\bf #1} }

%% file: sec/0_abstract.tex
\begin{abstract}
We introduce ECLAIR (\textbf{E}xtended \textbf{C}lassification of \textbf{L}idar for \textbf{AI R}ecognition), a new outdoor large-scale aerial LiDAR dataset designed specifically for advancing research in point cloud semantic segmentation. As the most extensive and diverse collection of its kind to date, the dataset covers a total area of 10$\text{km}^2$ with close to 600 million points and features eleven distinct object categories. To guarantee the dataset's quality and utility, we have thoroughly curated the point labels through an internal team of experts, ensuring accuracy and consistency in semantic labeling. The dataset is engineered to move forward the fields of 3D urban modeling, scene understanding, and utility infrastructure management by presenting new challenges and potential applications. As a benchmark, we report qualitative and quantitative analysis of a voxel-based point cloud segmentation approach based on the Minkowski Engine. We release the dataset as open-source and it can be accessed at \url{https://github.com/SharperShape/eclair-dataset}

\end{abstract}

%% file: sec/1_intro.tex
\vspace{-3mm}
\section{Introduction}\label{sec:intro}
Recent breakthroughs in the field of deep learning \cite{touvron2023llama, segmentanything, clip} are attributed to factors such as the availability of vast and extensive datasets and have enabled models to generalize effectively across diverse applications. However, such progress has not been mirrored in the domain of 3D LiDAR. For instance, the DALES dataset~\cite{dales-dataset}, comprising forty scenes, amounts to a few gigabytes. In contrast, the CommonCrawl dataset, one of the largest in Natural Language Processing (NLP) utilized by the LLaMA model \cite{touvron2023llama}, spans approximately 6 petabytes. Similarly in Computer Vision, the SegmentAnything\cite{segmentanything} dataset occupies 11.3 terabytes of storage. The inherent nature of 3D datasets, with aspects such as the dimensionality and point density, also contributes to this disparity. These characteristics pose unique challenges for their collection, labeling, and management. This paper aims to bridge this gap with the aspiration of increasing the availability of point cloud dataset with a more extensive and rich dataset comparable in size to DALES, and facilitate further research into the deep learning models and their quality.

\input{tables/datasets_comparison}

Outdoor 3D scene understanding is fundamental to many applications in computer vision, including autonomous driving, robotics, Augmented and Virtual Reality (AR / VR)~\cite{guzov2021human,sarlin2022lamar}. The last several years, modern machine learning techniques have advanced state-of-the-art scene understanding algorithms for object detection, depth estimation, semantic and instance segmentation, 3D reconstruction, and more. Most of these approaches are enabled through a diverse set of real and synthetic RGB-(D) datasets~\cite{s3-dis,kitti,sun-rgbd}. 

The demand for diverse and accurately annotated datasets captured at a large scale is becoming more critical in point cloud semantic segmentation techniques. These machine learning-based techniques are instrumental across a multitude of applications ranging from autonomous driving to urban planning. At present, existing datasets for point cloud semantic segmentation exhibit a significant inclination towards scenarios predominantly related to autonomous vehicles~\cite{semantic-kitti-dataset,semanticposs-dataset,nuscenes-dataset,lyft-dataset,waymo-dataset} that use Mobile Laser Scanning (MLS) or Terrestrial Laser Scanning (TLS) systems to collect the data. While these datasets have advanced perception systems for self-driving cars, their scope, largely confined to vehicle-centric perspectives, introduces a notable gap in the diversity and coverage environments. This limitation particularly overlooks the potential of aerially captured data. Other datasets~\cite{scannet,scannet200,yeshwanthliu2023scannetpp,s3-dis} provide detailed scans of interiors for tasks such as object recognition, semantic segmentation, and novel view synthesis. While these datasets have been instrumental in advancing indoor mapping and navigation systems, they are inherently limited in their applicability to outdoor, large-scale environments due to their specific focus on indoor spaces.

Airborne LiDAR Scanning (ALS) systems generate point cloud data that significantly differ from those from self-driving and indoor datasets. The orientation of ALS sensors is typically close to a nadir view, leading to distinct occlusions compared to ground-based scanning systems. Aerial LiDAR data collection is often more costly than mobile LiDAR due to expenses associated with aerial flights~\cite{isprs-dataset,dublin-city-dataset,dales-dataset}. To address the high costs associated with hardware,~\cite{campus-3d-dataset,sensat-urban-dataset,swiss-3d-dataset} propose generating point clouds from high-quality aerial images captured by a UAV-based mapping system using photogrammetry. Although cost-efficient, the quality of the reconstructed point cloud significantly depends on the discriminative performance of local image descriptors that can struggle to handle different lighting and weather conditions. The ALS systems provide advantages such as providing a more uniform density of point clouds and covering a broader area coverage during data collection. Moreover, it enables data collection in areas where terrestrial travel is challenging. These unique features make ALS ideal for urban planning and surveying applications.

The dataset we propose, named ECLAIR (\textbf{E}xtended \textbf{C}lassification of \textbf{L}idar for \textbf{AI R}ecognition), consists of a large-scale point cloud collected from a region in the city of Espoo, Finland. It covers a contiguous area of more than 10 square kilometers consisting of more than half a billion points captured by a long-range high-accuracy LiDAR. The focus of the data capture has been to cover the electrical transmission lines; consequently, the point clouds follow this network. A comparison of ECLAIR with some of the existing point cloud datasets is presented in~\cref{tab:dataset_comparison}. In addition to the raw data, we provide accurate ground truth and pseudo labels, and demonstrate their usability in a downstream supervised learning task: point cloud semantic segmentation. In contrast to DALES~\cite{dales-dataset}, ECLAIR further uses high-resolution nadir images to provide colorized point clouds. We describe the dataset capturing pipeline as well as the point cloud colorization process in~\cref{sec:dataset}. Along with color, the dataset also includes intensity, the return number, and the number of returns as features. Lastly, the proposed dataset not only shares similarities with existing datasets but also introduces unique distributions that, when combined with other datasets, facilitate large-scale generalized representational learning~\cite{ppt}.

In summary, we make the following contributions: 1) We introduce \textit{ECLAIR}, a new outdoor, large-scale aerial LiDAR dataset with point-wise semantic annotations; 2) The proposed dataset enables training and benchmarking point cloud segmentation approaches on large-scale, real-world scenes captured by a high-quality aerial LiDAR; 3) We thoroughly evaluate one of the existing voxel-based point cloud semantic segmentation approaches (\ie, the Minkowski Engine~\cite{choy20194d_minkowski}) on the proposed dataset and discuss quantitative results.

%% file: tables/datasets_comparison.tex
\begin{table*}[ht!] %
\scriptsize
\centering %
\resizebox{\textwidth}{!}{%
\centering
\begin{tabular}{lccccccc} %
\toprule %

Dataset & Category & Year & Spatial Size, $m$ / Area, $m^2$ & \# Classes & \# Points & \# RGB & Sensor \\ %
\midrule %
S3DIS~\cite{s3-dis} & \multirow{3}{*}{Indoor scene-level} & 2017 & $6 \times 10^3 \: m^2$ & 13 & 273M & \greencmark & Matterport \\
ScanNet~\cite{scannet} & & 2017 & $1.1 \times 10^5 \: m^2$ & 20 & 242M & \greencmark & RGB-D \\
ScanNet++~\cite{yeshwanthliu2023scannetpp} & & 2023 & $1.5 \times 10^4 \: m^2$ & 1000 & - & \greencmark & RGB-D \\
\midrule
Semantic3D~\cite{semantic3D} & \multirow{4}{*}{Outdoor road-level} & 2017 & - & 8 & 4000M & \greencmark & TLS \\
SemanticKITTI~\cite{semantic-kitti-dataset} &  & 2019 & $39.2 \times 10^3 \: m^2$ & 25 & 4549M & \redxmark & MLS \\
Toronto-3D~\cite{toronto-3d-dataset} &  & 2020 & $1 \times 10^3 \: m^2$ & 8 & 78.3M & \greencmark & MLS \\
SemanticPOSS~\cite{semanticposs-dataset} & & 2020 & - & 14 & 216M & \redxmark & MLS \\
\midrule
ISPRS~\cite{isprs-dataset} & \multirow{6}{*}{Aerial urban-level} & 2012 & - & 9 & 1.2M & \redxmark & ALS \\
DublinCity~\cite{dublin-city-dataset} &  & 2019 & $2 \times 10^6 \: m^2$ & 13 & 260M & \redxmark & ALS \\
Campus3D~\cite{campus-3d-dataset} &  & 2020 & $1.58 \times 10^6 \: m^2$ & 24 & 937.1M & \greencmark & P \\
SensatUrban~\cite{sensat-urban-dataset} &  & 2020 & $7.64 \times 10^6 \: m^2$ & 13 & 2847M & \greencmark & P \\
Swiss3DCities~\cite{swiss-3d-dataset} &  & 2020 & $2.7 \times 10^6 \: m^2$ & 5 & 226M & \greencmark & P \\
DALES~\cite{dales-dataset} &  & 2020 & $10 \times 10^6 \: m^2$ & 8 & 505M & \redxmark & ALS \\
\textbf{ECLAIR (ours)}&  & 2024 & $10.3 \times 10^6 \: m^2$ & 11 & 582M & \greencmark & ALS \\
\bottomrule %
\end{tabular}
}
\caption{\textbf{Comparison of datasets. } We compare existing datasets in terms of area of coverage, point density, and the sensor type. While the coverage area in DALES is similar to ours, the proposed dataset has more semantic classes and additionally provides colorized point clouds. Similar to~\cite{sensat-urban-dataset}, we use the following notation: MLS - Mobile Laser Scanning system; TLS - Terrestrial Laser Scanning system; ALS - Aerial Laser Scanning system; P - photogrammetry. }
\label{tab:dataset_comparison}
\end{table*}

%% file: sec/2_related_work.tex
\section{Related Work}\label{sec:related-work}
Deep learning approaches for 3D semantic understanding require diverse, large-scale datasets in order to generalize to new scenes. Here we give a brief introduction to existing datasets, compare them with ECLAIR, and provide an overview of current methods for point cloud semantic segmentation.

\subsection{Semantic Understanding of 3D Areas}
Existing datasets for large-scale point cloud segmentation can be widely categorized into three groups: indoor scene-level 3D datasets, outdoor road-level point cloud datasets, and urban-level aerial LiDAR datasets.

\boldparagraph{Indoor scene-level 3D datasets.} Early datasets in this category, such as SUN RGB-D~\cite{sun-rgbd}, NYUv2~\cite{nyu-v2}, and S3DIS~\cite{s3-dis}, represent RGB-D sequences captured by short-range depth scanners with low resolution and limited semantic annotations. Other datasets~\cite{scannet, scannet200, matterport3D} provide annotation at scale, but the performance on long-tail classes is limited by the resolution of ground truth geometry from laser scans. ARKitScenes~\cite{dehghan2021arkitscenes} and ScanNet++~\cite{yeshwanthliu2023scannetpp} address this limitation by incorporating both RGB images and high-resolution 3D scene geometry captured by lasers. They provide sparse (bounding boxes) and dense semantic annotations respectively.

\boldparagraph{Outdoor road-level 3D data.} This group of datasets is related to autonomous driving applications in which the data is captured by a LiDAR scanner together with RGB cameras mounted on a vehicle~\cite{semantic-kitti-dataset,paris-lille-dataset,semanticposs-dataset,nuscenes-dataset,astar-3d,waymo-dataset,toronto-3d-dataset,argoverse-dataset}. The mobile LiDAR datasets, with their low-angle perspective and emphasis on driving-related segmentation tasks, often result in occlusions inside the point clouds, \eg, missing roofs of buildings. While these datasets fulfill their primary purpose, they fall short for use in other domains, such as public utility asset management and urban planning.

\boldparagraph{Urban-level aerial datasets.} These datasets are pivotal for advancing research and applications in the fields of remote sensing, environmental monitoring, and autonomous navigation. They have primarily been obtained by aerial LiDARs~\cite{dales-dataset,dublin-city-dataset,lasdu-dataset,isprs-dataset} or by using photogrammetry~\cite{sensat-urban-dataset,swiss-3d-dataset,campus-3d-dataset}. In contrast to DALES~\cite{dales-dataset}, ECLAIR provides colorized, large-scale point clouds including high-resolution 3D geometry along with accurate semantic labels and the number of LiDAR returns for each point.

\subsection{3D Semantic Learning}
In general, deep learning based, point cloud semantic segmentation methods fall into three main categories based on their approach to modeling point clouds: projection-based, voxel-based, and point-based methods. Projection-based strategies convert 3D points onto different image planes, leveraging 2D CNN architectures to extract features~\cite{point-pillars,pointcnn,cenet}. Voxel-based methods~\cite{voxnet,choy20194d_minkowski,o-cnn}, on the other hand, turn point clouds into uniform voxel grids, making 3D convolution operations more manageable and improving their efficiency with sparse convolution techniques. In contrast, point-based approaches deal with point clouds directly with a notable recent trend towards adopting transformer-based architectures~\cite{zhao2021point-transformer,wu2022point-transformer-v2,wu2024-point-transformer-v3,robert2023spt}. In this work, we thoroughly evaluate one of the voxel-based approaches, using the Minkowski Engine~\cite{choy20194d_minkowski}, on the ECLAIR dataset and identify a number of key challenges revealed by our dataset.

%% file: sec/3_dataset_method.tex
\vspace{-2mm}
\section{ECLAIR: Dataset Creation}\label{sec:dataset}
This dataset was created through a multi-step process. In~\cref{ssec:data-capture}, data capture details the sensors and parameters employed in acquisition. Subsequent data processing, described in~\cref{ssec:data-process}, involved necessary transformations and preparation for analysis. Class specifications, crucial for semantic analysis, are defined in~\cref{ssec:class-specs}. Data curation included a manual examination step, outlined in~\cref{ssec:data-quality-control}. Finally, data visualization tools were utilized to illustrate the data and accelerate quality control, as discussed in~\cref{ssec:data-viz}.

\subsection{Data Capture}\label{ssec:data-capture}
The data presented are captured using our proprietary sensor system built in-house by Sharper Shape and mounted on a helicopter.
It is a single lightweight multi-sensor system, capable of collecting the data required for utility inspection and analysis via helicopter. It is equipped with the following hardware: a) Long-range, survey grade, high accuracy LiDAR coupled with highly precise GNSS and FOG IMU sensors; b) High resolution RGB cameras capturing oblique and ortho imagery; c) Push broom hyperspectral cameras providing a broad spectrum of wavelength along the flight path; and, d) Ambient temperature and humidity sensors.

All sensors were calibrated according to manufacturer specifications before data processing. The system has the capability to attach more sensors such as 4-band, ultraviolet sensors, etc. It has also been tested under various conditions and has been a reliable system enabling data capture across multiple projects under Sharper Shape.
The data was collected at a flight height of 100 meters and speed of 40 knots. With 600 PRR (Pulse Repetition Rate) and 234.5 lines per second, the LiDAR has a point density of 50 $\text{points}/\text{m}^2$ and a swath width of 328 meters. The LiDAR was calibrated using multiple overlap captures with a standard deviation error of less than 2 cm.

\subsection{Data Processing}\label{ssec:data-process}
The data preparation process starts with the utilization of the RiProcess software to convert raw files into the LAZ file format. Subsequently, the LAZ files are fed into the tiling pipeline in which the point clouds are partitioned into smaller $100\times100$ m tiles. This is facilitated by a tool developed internally by Sharper Shape to enhance data manageability.

Following tiling, the point clouds undergo a colorization process. For each tile, relevant images are selected by identifying areas of overlap. Subsequently, utilizing the camera's parameters, each point from the point cloud is projected from 3D onto 2D space. Color information from the corresponding projected location within the selected image is then extracted and applied to the point. In instances where multiple images contribute to a specific point, color averaging is performed across these images. However, this averaging mechanism may introduce inaccuracies, particularly in the coloring of thin structures such as powerlines. It is also worth noting that the coverage of images is lower than that of the point cloud, and hence approximately 20\% of the points lack color information.

Following colorization, the data undergoes point cloud segmentation utilizing a deep learning model. The employed proprietary model produces a total of 30 classes with a predominant focus on classes pertinent to electrical infrastructure. These classes are subsequently remapped to a set of 11 classes that are presented here. This remapping yields a notable reduction in classification errors, allowing the remapped version of the dataset to be effectively utilized for model training with only minimal manual intervention.

This comprehensive data preparation pipeline involves sequential processes aimed at converting, tiling, enhancing, and classifying point cloud data for various downstream applications including vectorization, vegetation encroachment analysis, etc. The coordinate system of the data is WGS 84 / UTM zone 35N (EPSG:32635).

\begin{figure}[t!]
    \centering
    \includegraphics[width=1.0\linewidth]{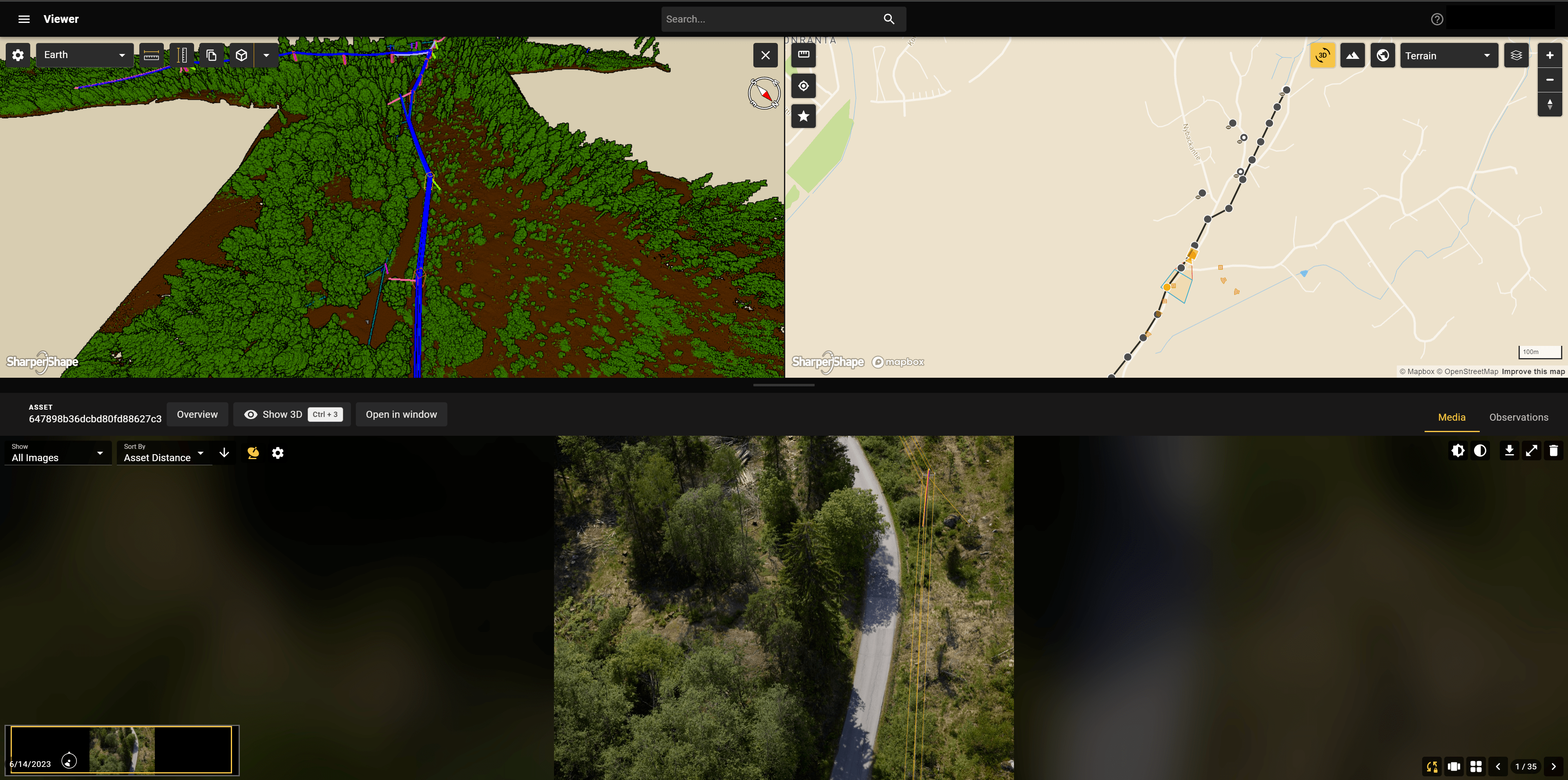}
    \caption{\textbf{Point cloud visualization}. CORE viewer combines a view of point clouds colored based on classifications, a map view, and image view. The point clouds and images also show the vector data of objects in 3D and 2D, respectively.}
    \label{fig:core-viewer}
\end{figure}

\subsection{Class Specifications}\label{ssec:class-specs}
We identify the following list of semantic classes with each corresponding label ID provided in parentheses:

\noindent \textit{Ground} (2) : All points representing the Earth's surface, including, soil, pavement, roads, and the bottom of water bodies.

\noindent \textit{Vegetation} (3) : All points representing organic plant life, ranging from trees, low shrubs, and tall grass of all heights.

\noindent \textit{Buildings} (4) : Man-made structures characterized by roofs and walls, encompassing houses, factories, and sheds.

\noindent \textit{Noise} (5) : Sporadic points suspended in air or underground.

\noindent \textit{Transmission Wires} (6) : High-voltage wires for long-distance transmission from power plants to substations. Either directly connected to transmission towers or poles. Also includes transmission ground wires.

\noindent \textit{Distribution Wires} (7) : Lower-voltage overhead distribution wires distributing electricity from substation to end users. Includes span guy wires and communication wires.

\noindent \textit{Poles} (8) : Utility poles used to support different types of wires or electroliers. These can include poles with either transmission or distribution wires. Down guy wires, crossarms and transformers are also included in this class.

\noindent \textit{Transmission Towers} (9) : Large structures supporting transmission wires with the distinct characterisation of steel lattices and cross beams.

\noindent \textit{Fence} (10) : Barriers, railing, or other upright structure, typically of wood or wire, enclosing an area of ground.

\noindent \textit{Vehicle} (11) : All wheeled vehicles that can be driven.

\noindent \textit{Unassigned} (1) : This category serves as a catch-all for non-subject points. Anything that is not on the class list is classified as Unassigned. These include wooden pallets, trash, structures not large or strong enough to put under buildings (tents, boulders, etc.), and house antennas.

\subsection{Data Quality Control}\label{ssec:data-quality-control}
The dataset classifications are derived from fully automated processes which may contain errors. To discern accurate classifications within the dataset, manual verification of tiles was conducted by Sharper Shape's internal data curation team of annotation experts in the power utility domain. The dataset is bifurcated into two primary categories: ``Ground Truth'' and ``Pseudo-Labels''. Within each tile, if misclassifications for a particular object class (excluding \textit{Ground}, \textit{Vegetation}, \textit{Unassigned}, and \textit{Noise}) exceed 10 points, the tile is categorized as a pseudo-label. Conversely, tiles devoid of misclassifications or with misclassifications totaling fewer than 10 points for object classes are allocated to the ``Ground Truth'' category. Overall, out of the 1246 tiles in the dataset, 624 are classified as ground truth, and 622 are categorized as pseudo labels.

\subsection{Data Visualization}\label{ssec:data-viz}
In order to facilitate quality control and to easily visualize the data, a software platform named Sharper CORE was used. This has been developed internally by Sharper Shape. It enables the use of geographical information with point clouds and images, allowing for sensor fusion. This enables comprehensive data analysis to make sure the quality aligns well with the expectation during quality control tasks.

\begin{figure}[t!]
    \centering
    \includegraphics[width=\linewidth]{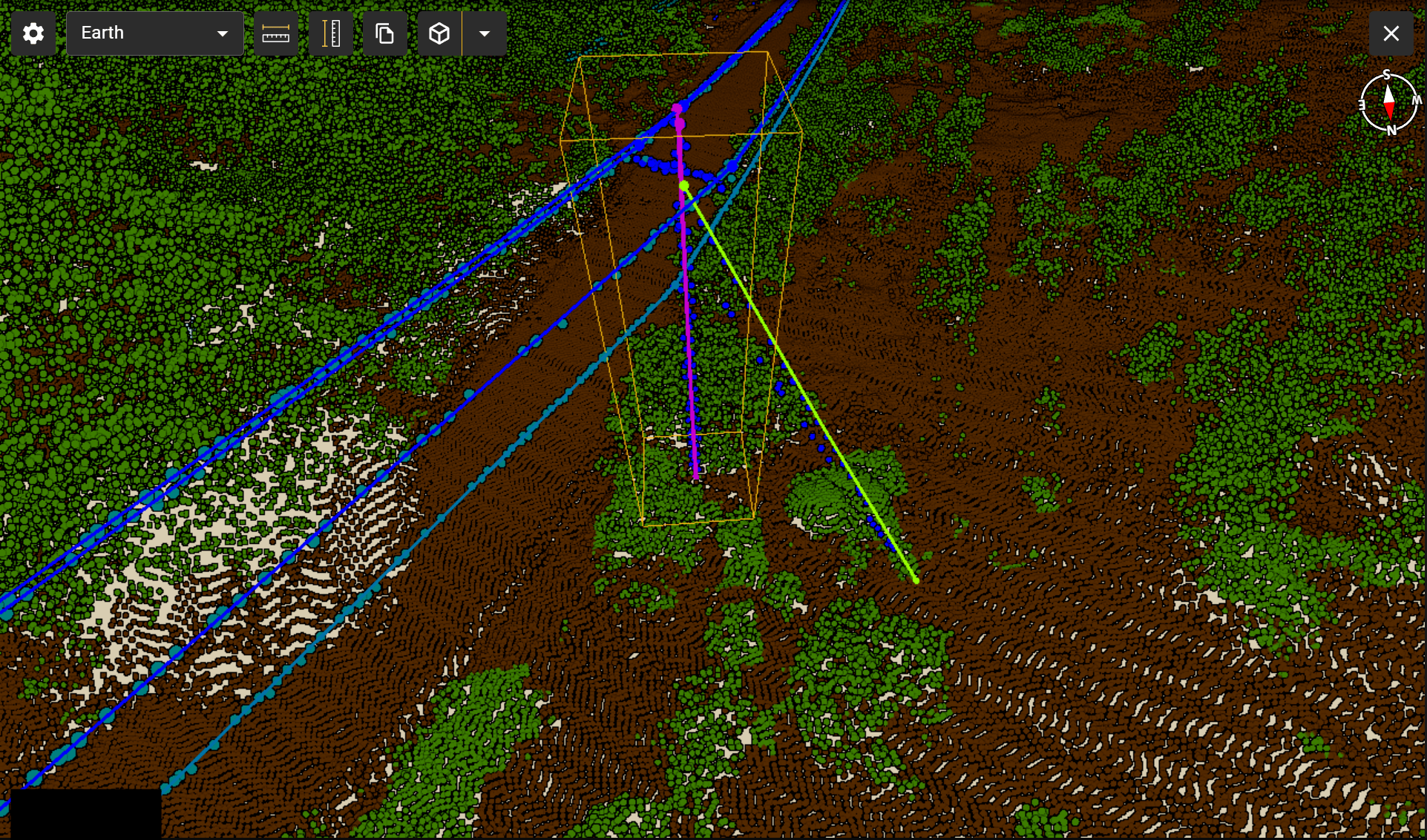}
    \caption{\textbf{Point cloud inspection.} 3D Point Cloud Navigation/Editing View provided by CORE.}
    \label{fig:3d-vectors}
\end{figure}

\cref{fig:core-viewer} shows a screenshot from the CORE web software, showcasing various views available within the interface. The illustration highlights the integration of data from different modalities into a unified view, offering extensive contextual information during quality control tasks. In addition to visualizing point clouds and images, CORE also supports visualizing vector overlays which help with assessing quality of object classes such as power lines, poles, towers, etc. as illustrated in  %
\cref{fig:3d-vectors}. This context is valuable since we no longer have to navigate every portion of point cloud data to evaluate our deep learning models and data.

\begin{figure}[t!]
    \centering
    \includegraphics[width=\linewidth]{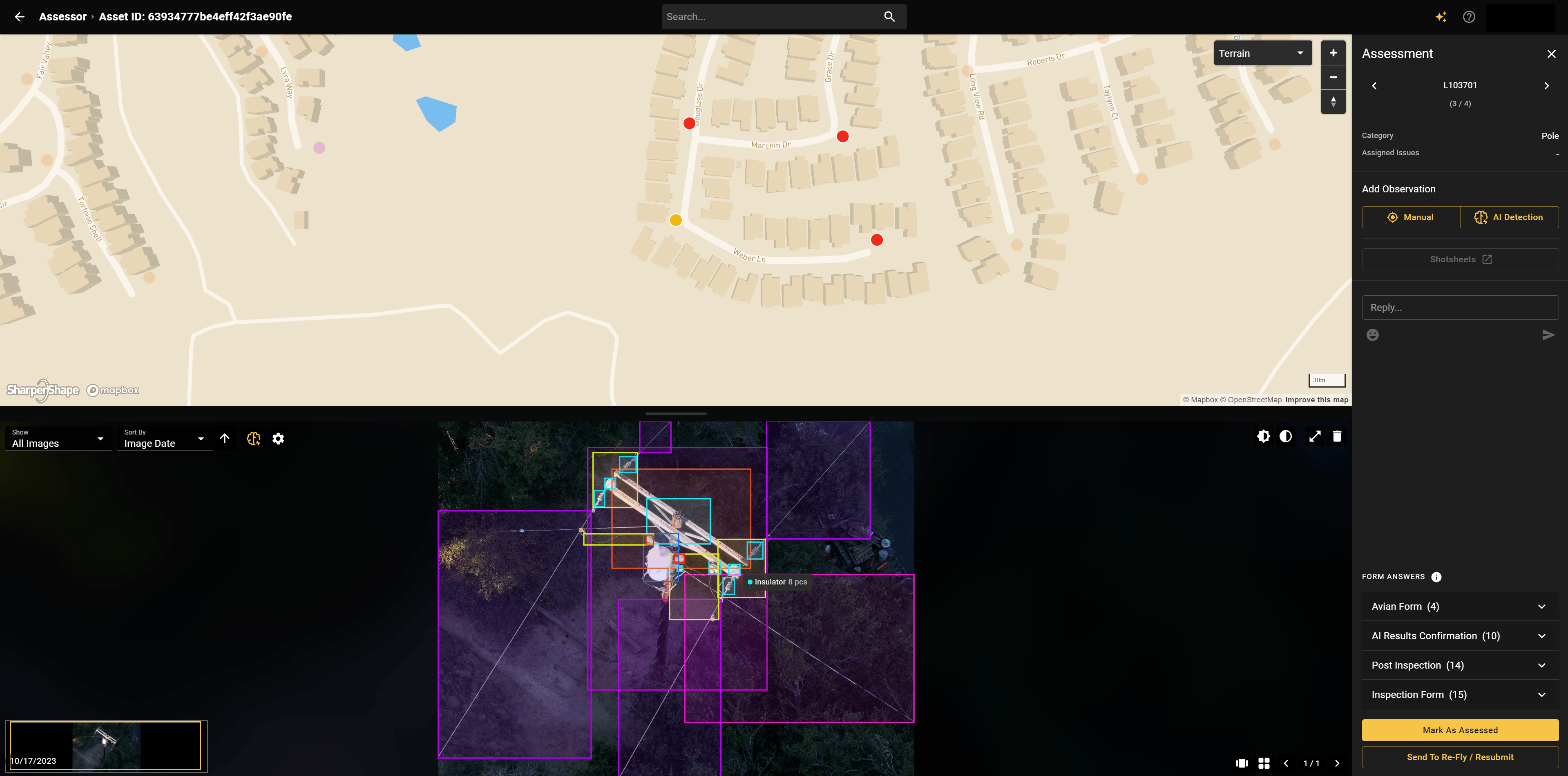}
    \caption{\textbf{Component inventory.} Detection and inventory of specific components in CORE based on multimodal data inputs.}
    \label{fig:aici}
\end{figure}

CORE is built on top of specialized GIS databases which store information about the vector models that can be used to do further analysis and reporting. This allows for in-depth component inventory and environmental analysis in a structured and meaningful way so that data can be queried to build specific datasets. An example of inventory is shown in ~\cref{fig:aici}. In addition, CORE scales with large amounts of data seamlessly and allows custom rendering settings and cloud-based data serving. As a result, numerous individuals with limited technical proficiency in point cloud data were able to participate in the quality control process, collaborating simultaneously online with the aid of basic instructions.

%% file: sec/4_experiments.tex
\input{tables/ablations}

\section{Experimental Evaluation}\label{sec:experiments}
We first present statistics of the proposed dataset in~\cref{ssec:dataset-statistics} and discuss evaluation metrics in~\cref{ssec:metrics}. Next, we analyze different design choices and perform ablation studies of the Minkowski Engine-based baseline network with 4 different architectures and various sets of point features. We also provide details of loss functions that can efficiently handle class imbalance and investigate the influence of pseudo labels on semantic segmentation performance in~\cref{ssec:ablation}. Finally, we provide quantitative and qualitative results of the baseline model on the proposed dataset.

\subsection{Statistics of ECLAIR}\label{ssec:dataset-statistics}
The proposed dataset comprises 1246 tiles, each covering an area of $100\times100$ square meters. To ensure a robust evaluation framework, the dataset is divided into train, validation, and test splits following the standard proportions of 70\%, 10\%, and 20\% respectively. The validation and test splits consist only of the ground truth tiles to ensure that the metrics generated are reliable and consistent.

\begin{figure}[t!]
    \centering
    \includegraphics[width=\linewidth]{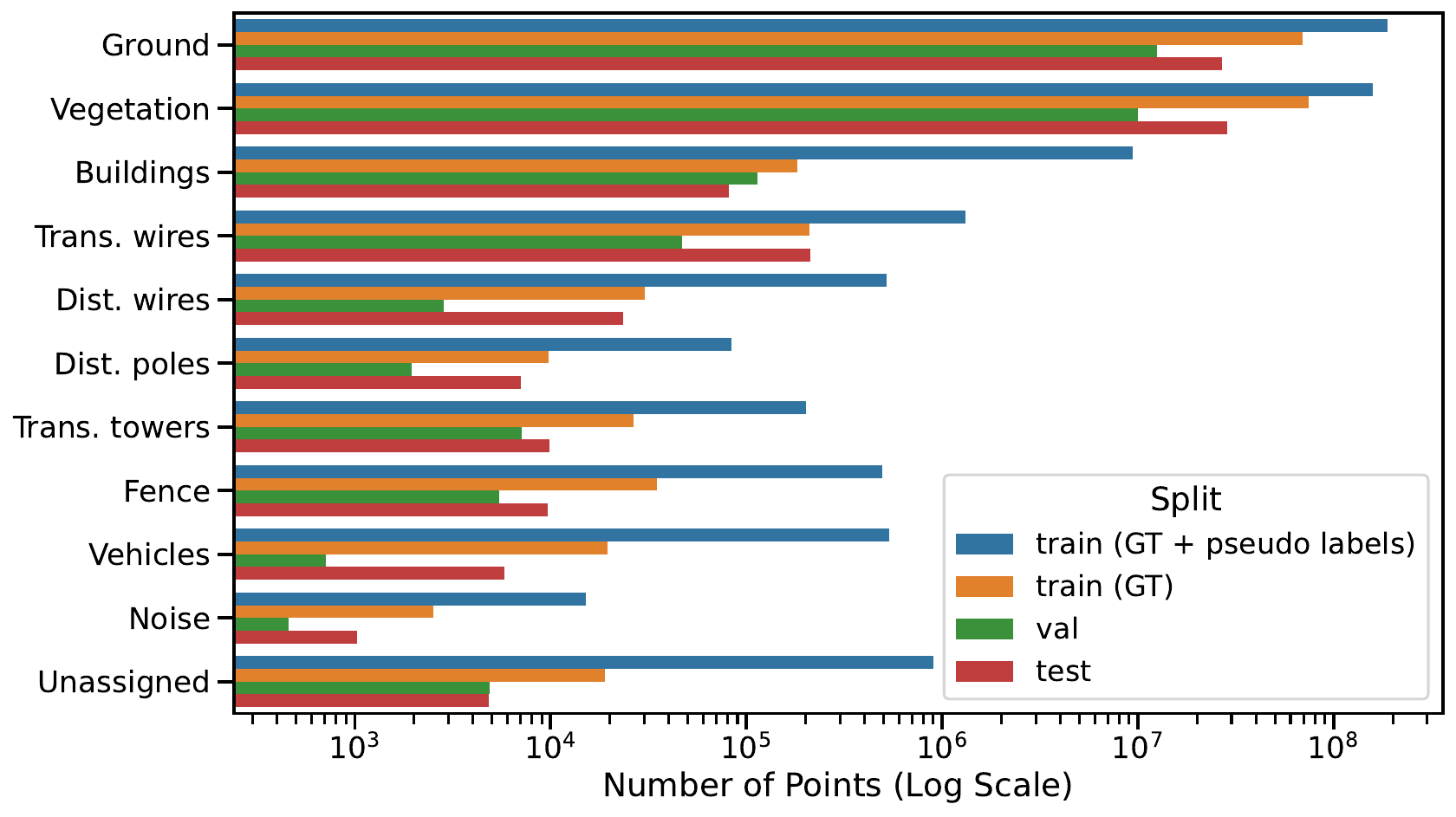}
    \vspace{-7mm}
    \caption{\textbf{The distribution of semantic classes.} We report the total number of points for each semantic category showing a high imbalance of the proposed dataset (note the logarithmic scale for the horizontal axis).   }
    \label{fig:dataset-stats}
    
\end{figure}

\cref{fig:dataset-stats} provides statistics over all points of the proposed dataset revealing a significant imbalance across the semantic labels.  Predominant categories such as Ground, Vegetation, and Buildings are overrepresented forming the majority of the dataset. In contrast, critical but less frequent categories (\eg, Transmission Towers, Distribution Wires, Poles, and Vehicles) account for less than 1\% of the total number of points, underscoring a challenge in achieving balanced representation. This imbalance reflects real-world conditions, presenting an opportunity to test the robustness and generalizability of point cloud classification models under skewed distribution scenarios. In addition to the carefully curated tiles (\texttt{GT} in~\cref{fig:dataset-stats}), we also release a subset with pseudo labels generated by our proprietary point cloud classification model (\texttt{pseudo labels} in~\cref{fig:dataset-stats}). Although having imperfect semantic labels, this dataset improves the model's generalization performance leading to better segmentation results (\cf~\cref{ssec:ablation}).

\vspace{-2mm}
\subsection{Metrics}\label{ssec:metrics}
The F1 score and Intersection over Union (IoU) are both metrics used to evaluate the performance of semantic segmentation models, including those applied to point clouds. The F1 score is the harmonic mean of precision and recall. Precision measures the correctness of the positive predictions made by the model, while recall measures the model's ability to detect all actual positives.
The F1 score is beneficial when the balance between precision and recall is required, especially when there is an uneven class distribution. It ensures that a model is not simply predicting the majority class. In contrast, IoU may be less informative in scenarios in which class imbalance affects the model's performance as the score primarily focuses on the spatial accuracy of the segmentation and not on the model's ability to detect rare classes. Therefore, in scenarios when it is essential to both identify every instance of a given class (recall) and ensure the accuracy of these detected instances (precision), particularly in context of significant class imbalance (\cf~\cref{ssec:dataset-statistics}), we employ the F1 score as a key metric for evaluating the performance of semantic segmentation models. To be consistent with other works~\cite{dales-dataset,campus-3d-dataset,semanticposs-dataset}, we also report the IoU.

\vspace{-2mm}
\subsection{Ablation Studies}\label{ssec:ablation}
In this section, we conduct several ablation studies to examine the impact of various factors on segmentation performance. These factors include different architectural design choices using the Minkowski Engine~\cite{choy20194d_minkowski}, the effect of diverse point features and objective functions, and the quality of semantic ground-truth labels.

\boldparagraph{Point features.} We consider the following features: a) Intensity $f_\text{int}$; b) Return data $f_\text{return}$: it includes the LiDAR return number and the number of returns; c) Color $f_\text{color}$; d) a combination of Intensity and Return data $f_\text{int}+f_\text{return}$; e) a combination of Intensity and Color $f_\text{int}+f_\text{color}$; f) a combination of Return data and Color $f_\text{return}+f_\text{color}$; g) all the features combined, \ie $f_\text{int}+f_\text{return}+f_\text{color}$. The results presented in~\cref{stab:ablation-features} show that combining intensity and return data achieves the best performance. The color feature seems quite powerful and can also improve semantic segmentation results for important classes, such as \textit{Fence}.

\boldparagraph{Loss functions.} Fundamentally, urban areas typically exhibit a highly skewed distribution of categories with a few dominant classes such as vegetation and ground occupying the majority of points, while smaller, yet critical, categories such as wires constitute a tiny fraction of points. The highly imbalanced distribution presents a major challenge from the ECLAIR dataset for accurate semantic segmentation (\cf~\cref{ssec:dataset-statistics}). To address this issue, adopting more advanced loss functions is a common strategy~\cite{focal_loss,berman2018lovasz,seesaw_loss}. We assess the efficacy of three off-the-shelf available loss functions, using the Minkowski Engine~\cite{choy20194d_minkowski} as a baseline model. The evaluated objective functions include: cross-entropy $\mathcal{L}_\text{ce}$, weighted cross-entropy based on inverse frequency $\mathcal{L}_\text{iwce}$, and the focal loss $\mathcal{L}_\text{fl}$~\cite{focal_loss}. The quantitative comparison of the baseline model with different loss functions is presented in~\cref{stab:ablation-losses}. We observe that using Focal loss leads to the best results, indicating that it can efficiently handle rare classes, \eg, \textit{Distribution Poles}.

\begin{figure}[t!]
    \centering
    \includegraphics[width=\linewidth]{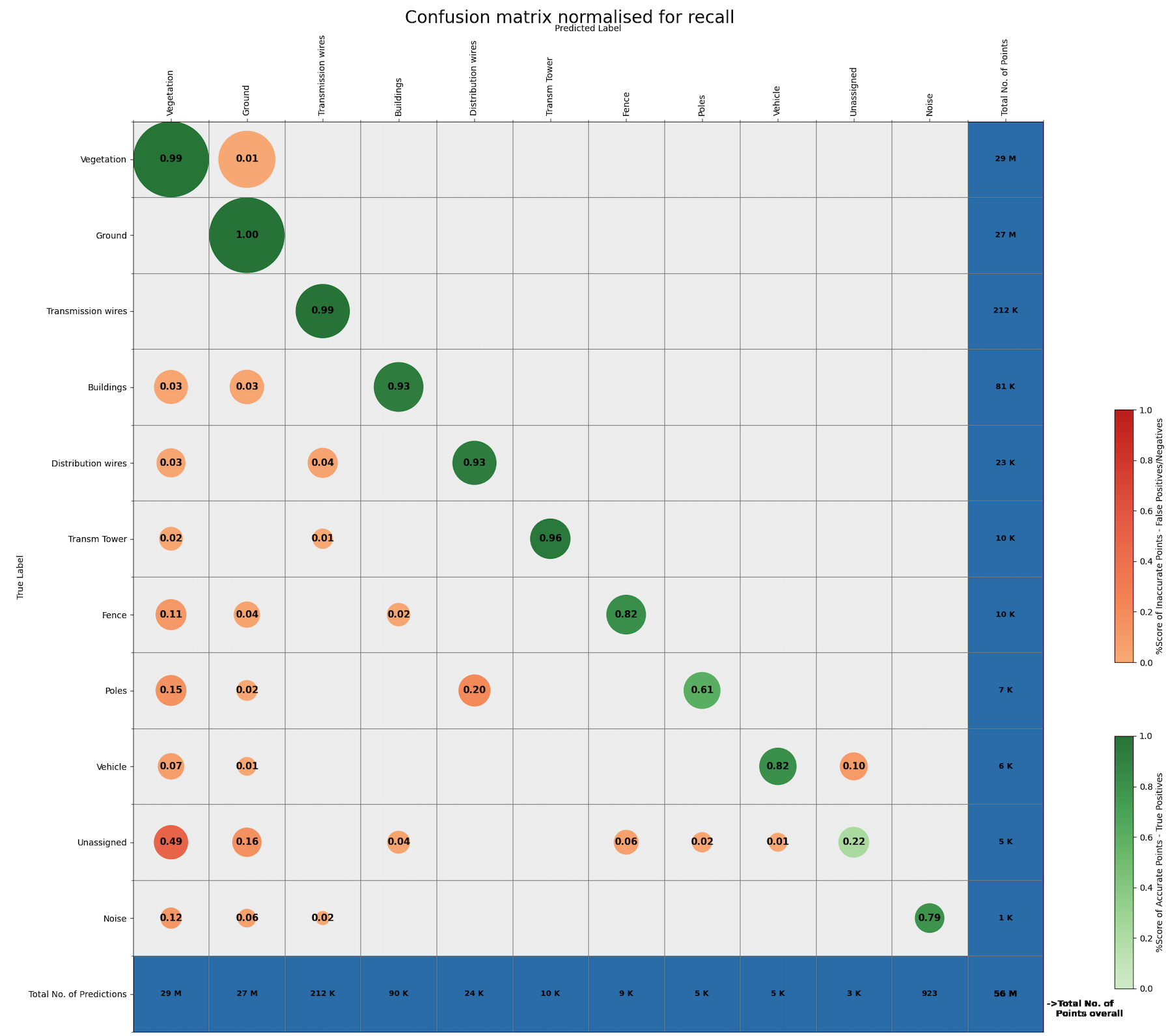}
    \caption{\textbf{Confusion matrix.} We report semantic segmentation results for our best model using a confusion matrix. Here, the size of circles corresponds to the total amount of points of each semantic class. The per-class F1 score is reported inside each circle. We find that the model performs well overall but falls short at segmenting rare, important classes, \eg, \textit{Poles}. Please zoom in to see the details.}
    \label{fig:res-confusion-matrix}
\end{figure}

\boldparagraph{Ground truth vs. pseudo labels.} We ran experiments with the two point cloud annotation strategies discussed in \cref{ssec:data-quality-control} and \cref{ssec:dataset-statistics}. As shown in~\cref{stab:ablation-annotations}, adding pseudo labels obtained by our proprietary point cloud segmentation approach leads to consistent improvements across all the classes compared to the manually verified ground truth data. Inaccurate labels introduce a form of noise into the training process, which can help the model learn to generalize better by forcing it to learn from a broader range of examples than it might from a smaller, perfectly labeled dataset (\cf~\cref{fig:dataset-stats}: train (GT) vs. train (GT + pseudo labels)).

\begin{figure*}[t!]
    \centering
    \includegraphics[width=\textwidth]{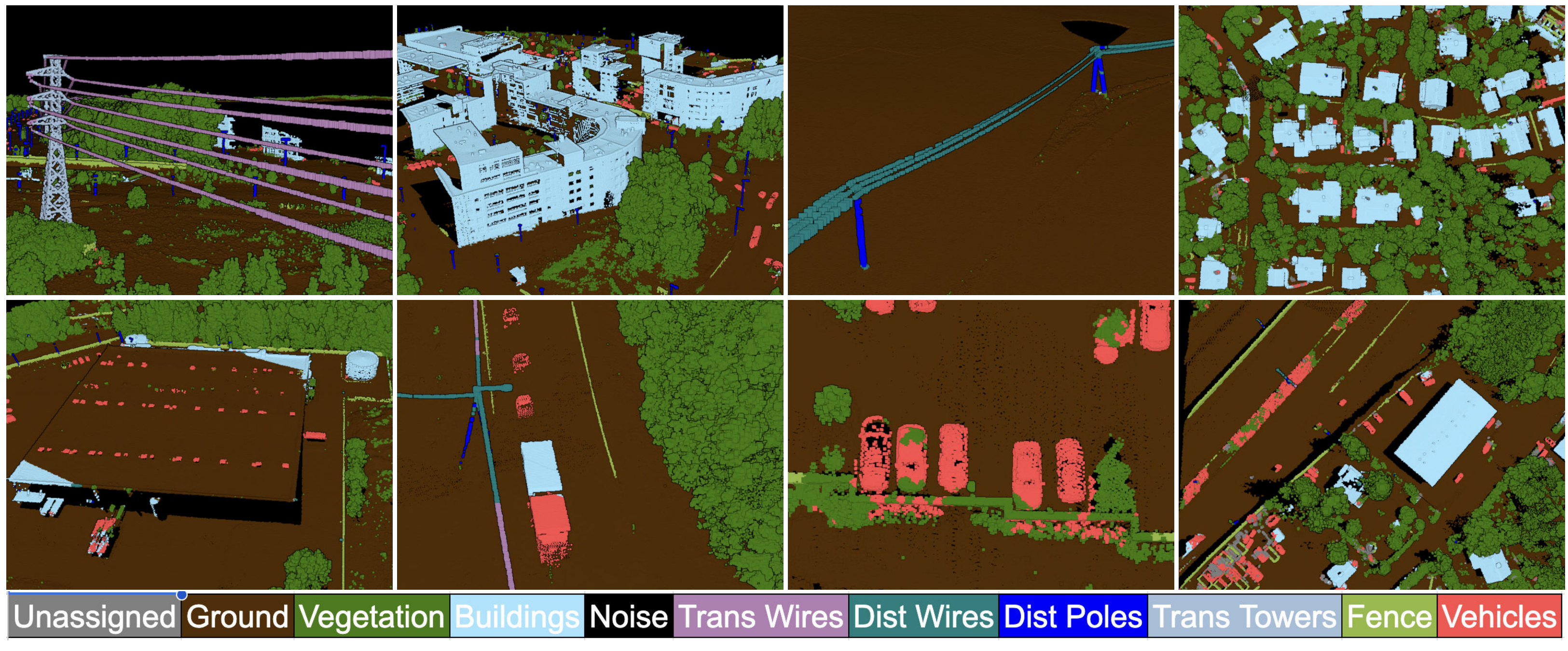}
    \caption{\textbf{Qualitative results of the best model.} We assess segmentation performance of the strongest baseline model on the test split of the proposed ECLAIR dataset. The top row demonstrates successful cases where the model performs well. In contrast, the bottom row highlights failure scenarios where the model produces noisy, inaccurate predictions.}
    \label{fig:inference-results}
\end{figure*}

\boldparagraph{Network architecture.} In order to evaluate the performance with different model capacities, we consider the Minkowski Engine~\cite{choy20194d_minkowski} with different number of layers. We list the architecture configurations and report corresponding semantic segmentation performance in~\cref{stab:ablation-architectures}.  Our experiments show that using the Res16UNet14C architecture improves overall segmentation quality achieving the best per-class performance.

\subsection{3D Semantic Understanding}

\boldparagraph{Technical details.} The model architecture utilized in this study is ResUNet \cite{resunet} implemented with the Minkowski Engine \cite{choy20194d_minkowski}. Conventional convolution layers are replaced with spatial convolution layers to accommodate point cloud data. Training of the model is conducted on AWS g5.12xlarge instances, each equipped with 4 NVIDIA A10G GPUs with 24 GB of vRAM per GPU. Given the memory-intensive nature of these networks, we maintain a batch size of 2 per GPU, resulting in an effective batch size of 8. Tiles are cropped to ensure a maximum size of $100\text{m} \times 100\text{m}$. Data augmentation techniques such as random coordinate scaling, random jitter, and random flip are applied, followed by normalization of coordinates before quantization. For features, RGB and intensity are scaled from 0 to 1 as floating-point values, and return number and number of returns are one-hot encoded before being fed into the network. The Adam optimizer~\cite{adam-optimizer} is employed with a learning rate set to 0.001, augmented by a step scheduler to adjust the learning rate every 10 steps. The voxel size is set to 0.05 with a normalisation factor of 10.0 applied to the coordinates before voxelisation.

\boldparagraph{Analysis.} According to the ablation study performed in~\cref{ssec:ablation} and reported in~\cref{tab:ablations}, we chose the Res16Unet14C architecture as a backbone of the Minkowski Engine network. The model was trained using both point features, \ie $f_{\text{int}}$+$f_{\text{return}}$ with Focal Loss~\cite{focal_loss} achieving macro F1 of $0.848$ and macro IoU of $77.35\%$. To further investigate segmentation performance of our baseline model, we provide a confusion matrix illustrated in~\cref{fig:res-confusion-matrix}. As can be seen, the model performs well successfully classifying the major categories, such as \textit{Ground}, \textit{Vegetation}, \textit{Transmission Wires}, and \textit{Buildings}, while classes such as \textit{Fence} and \textit{Poles} have very poor generalization scores. Similar to~\cite{dales-dataset,semanticposs-dataset}, we believe that the imbalanced distribution of semantic classes significantly affects the model's ability to generalize, as it mainly aligns with dominant classes while struggling to effectively capture the distinct characteristics of less representative but important classes. Qualitative results are illustrated in ~\cref{fig:inference-results}.

%% file: tables/ablations.tex
\begin{table*}[t!]
\centering
\begin{subtable}[t!]{\textwidth}
\scriptsize
\resizebox{\textwidth}{!}{%
\begin{tabular}{l|cc|ccccccccccc}
\toprule
 & \multicolumn{2}{c|}{macro} & \multicolumn{11}{c}{per-class F1 / IoU (\%)} \\
 {Features} & {F1} &  {IoU (\%)} & {Ground} & {Vegetation} & {Buildings} & {Noise} & {Trans. wires} & {Dist. wires} & {Dist. poles} & {Trans. towers} & {Fence} & {Vehicles} & {Unassigned}\\
\midrule
 $f_{\text{int}}$ & 0.841 & 76.72 & \textbf{0.99} / 98.23 & \textbf{0.99} / 98.32 & 0.92 / 85.87 & 0.81 / 68.09 & \textbf{0.99} / 98.86 & 0.90 / 83.19 & 0.73 / 57.50 & 0.94 / 90.10 & \textbf{0.85} / 73.88 & 0.86 / 76.35 & 0.23 / 13.58\\
 $f_{\text{return}}$ & 0.842 & 76.63 & \textbf{0.99} / 98.30 & \textbf{0.99} / 98.40 & 0.93 / 86.82 & \textbf{0.83} / \textbf{71.04} & \textbf{0.99} / 98.87 & 0.90 / 82.26 & 0.69 / 52.27 & \textbf{0.96} / \textbf{91.62} & 0.83 / 70.72 & 0.86 / 75.34 & 0.29 / 17.24\\
 $f_{\text{color}}$ & 0.838 & 75.69 & \textbf{0.99} / \textbf{98.33} & \textbf{0.99} / \textbf{98.43} & 0.92 / 86.43 & 0.81 / 68.61 & \textbf{0.99} / \textbf{99.20} & \textbf{0.92} / \textbf{85.65} & 0.69 / 52.86 & 0.94 / 88.08 & 0.84 / 72.65 & 0.77 / 61.94 & \textbf{0.34} / \textbf{20.46} \\
 $f_{\text{int}}$+$f_{\text{color}}$ & 0.828 & 74.80 & \textbf{0.99} / 98.32 & \textbf{0.99} / \textbf{98.43} & 0.92 / 85.43 & 0.81 / 67.75 & \textbf{0.99} / 98.80 & 0.90 / 82.13 & 0.68 / 51.02 & 0.95 / 90.01 & \textbf{0.85} / \textbf{74.50} & 0.76 / 74.50 & 0.26 / 15.14\\
 $f_{\text{return}}$+$f_{\text{color}}$ & 0.829 & 74.64 & \textbf{0.99} / 98.18 & \textbf{0.99} / 98.29 & 0.92 / 84.38 & 0.81 / 68.61 & \textbf{0.99} / 98.68 & 0.89 / 79.80 & 0.70 / 53.23 & 0.93 / 87.63 & 0.81 / 68.66 & 0.81 / 68.03 & 0.27 / 15.52\\
 $f_{\text{int}}$+$f_{\text{return}}$ & \textbf{0.848} & \textbf{77.35} &  \textbf{0.99} / 98.27 & \textbf{0.99} / 98.36 & \textbf{0.94} / \textbf{87.73} & 0.82 / 69.02 & \textbf{0.99} / 98.51 & 0.89 / 80.69 & \textbf{0.74} / 58.65 & 0.95 / 90.87 & 0.83 / 70.39 & \textbf{0.89} / \textbf{80.49} & 0.30 / 17.89 \\
 $f_{\text{int}}$+$f_{\text{return}}$+$f_{\text{color}}$ & 0.843 & 76.48 & \textbf{0.99} / 98.27 & \textbf{0.99} / 98.41 & 0.88 / 78.76 & \textbf{0.83} / 70.89 & \textbf{0.99} / 98.95 & \textbf{0.92} / 84.50 & \textbf{0.74} / \textbf{59.06} & 0.95 / 88.36 & \textbf{0.85} / 73.62 & 0.85 / 73.87 & 0.28 / 16.51\\
 
\bottomrule
\end{tabular}
}
\caption{Point features. We compare the three different point features: intensity $f_\text{int}$; the return number and the number of returns $f_\text{return}$; the color $f_\text{color}$ and their combinations (\cf~\cref{ssec:ablation}) and report macro and per-class F1 and IoU metrics. We observe that the combination of intensity and return features achieves the best segmentation results.}\label{stab:ablation-features}
\end{subtable}
\vspace{0.4mm}

\begin{subtable}[t!]{\textwidth}
\scriptsize
\resizebox{\textwidth}{!}{%
\begin{tabular}{l|cc|ccccccccccc}
\toprule
 & \multicolumn{2}{c|}{macro} & \multicolumn{11}{c}{per-class F1 / IoU (\%)} \\
 {Loss function} & {F1} &  {IoU (\%)} & {Ground} & {Vegetation} & {Buildings} & {Noise} & {Trans. wires} & {Dist. wires} & {Dist. poles} & {Trans. towers} & {Fence} & {Vehicles} & {Unassigned}\\
\midrule
 Minkowski+$\mathcal{L}_\text{ce}$ & 0.831 & 75.45 & \textbf{0.99} / \textbf{98.36} & \textbf{0.99} / \textbf{98.46} & \textbf{0.93} / \textbf{87.78} & 0.80 / 66.43 & \textbf{0.99} / 98.73 & 0.90 / 80.00 & 0.62 / 45.08 & 0.94 / \textbf{89.06} & \textbf{0.85} / \textbf{73.72} & \textbf{0.88} / \textbf{77.99} & 0.25 / 14.32 \\
 Minkowski+$\mathcal{L}_\text{iwce}$ & 0.726 & 63.70 & \textbf{0.99} / 97.94 & 0.98 / 98.05 & 0.84 / 72.48 & 0.30 / 17.78 & \textbf{0.99} / 98.85 & 0.92 / 85.84 & 0.50 / 32.89 & 0.91 / 82.61 & 0.60 / 43.30 & 0.75 / 60.27 & 0.19 / 10.72 \\
 Minkowski+$\mathcal{L}_\text{fl}$ & \textbf{0.843} & \textbf{76.48} & \textbf{0.99} / 98.27 & \textbf{0.99} / 98.41 & 0.88 / 78.76 & \textbf{0.83} / \textbf{70.89} & \textbf{0.99} / \textbf{98.95} & \textbf{0.92} / \textbf{84.50} & \textbf{0.74} / \textbf{59.06} & \textbf{0.95} / 88.36 & \textbf{0.85} / 73.62 & 0.85 / 73.87 & \textbf{0.28} / \textbf{16.51}\\
\bottomrule
\end{tabular}
}
\caption{Loss functions. We train the Minkowski Engine~\cite{choy20194d_minkowski} with different loss functions to handle class imbalance of the ECLAIR dataset. The following notation is used: $\mathcal{L}_\text{fl}$ - Focal loss; $\mathcal{L}_\text{ce}$ - Cross-Entropy loss; $\mathcal{L}_\text{iwce}$ - inverse weighted Cross-Entropy loss. }\label{stab:ablation-losses}
\end{subtable}
\vspace{0.4mm}

\begin{subtable}[t!]{\textwidth}
\small
\resizebox{\textwidth}{!}{%
\begin{tabular}{l|cc|ccccccccccc}
\toprule
 & \multicolumn{2}{c|}{macro} & \multicolumn{11}{c}{per-class F1 / IoU (\%)} \\
 {Training data} & {F1} &  {IoU (\%)} & {Ground} & {Vegetation} & {Buildings} & {Noise} & {Trans. wires} & {Dist. wires} & {Dist. poles} & {Trans. towers} & {Fence} & {Vehicles} & {Unassigned}\\
\midrule
 GT & 0.561 & 46.45 & 0.99 / 97.10 & 0.99 / 97.10 & 0.62 / 44.60 & 0.68 / 51.09 & 0.96 / 91.86 & 0.58 / 40.84 & 0.11 / 05.92 & 0.35 / 21.10 & 0.61 / 43.50 & 0.30 / 17.66 & 0.0 / 0.0   \\
Pseudo Labels & 0.842 & 76.35 & \textbf{0.99} / \textbf{98.27} & \textbf{0.99} / 98.36 & \textbf{0.95} / \textbf{91.02} & 0.80 / 66.55 & \textbf{0.99} / 98.27 & 0.87 / 76.88 & 0.69 / 52.71 & 0.94 / \textbf{88.99} & \textbf{0.85} / 73.52 & \textbf{0.87} / \textbf{76.62} & \textbf{0.31} / \textbf{18.59} \\
 GT+Pseudo Labels & \textbf{0.843} & \textbf{76.48} & \textbf{0.99} / \textbf{98.27} & \textbf{0.99} / \textbf{98.41} & 0.88 / 78.76 & \textbf{0.83} / \textbf{70.89} & \textbf{0.99} / \textbf{98.95} & \textbf{0.92} / \textbf{84.50} & \textbf{0.74} / \textbf{59.06} & \textbf{0.95} / 88.36 & \textbf{0.85} / \textbf{73.62} & 0.85 / 73.87 & 0.28 / 16.51\\

\bottomrule
\end{tabular}
}
\caption{Training data. Comparing different annotation strategies (\cf~\cref{ssec:data-quality-control}), we observe that using only carefully curated tiles (GT) leads to poor semantic segmentation results due to a lack of data for rare classes (\cf~\cref{ssec:dataset-statistics}). The pseudo labels combined with ground-truth significantly improve segmentation performance.}\label{stab:ablation-annotations}
\end{subtable}
\vspace{0.4mm}

\begin{subtable}[t!]{\textwidth}
\small
\resizebox{\textwidth}{!}{%
\begin{tabular}{l|cc|ccccccccccc}
\toprule
 & \multicolumn{2}{c|}{macro} & \multicolumn{11}{c}{per-class F1 / IoU (\%)} \\
 {Training data} & {F1} &  {IoU (\%)} & {Ground} & {Vegetation} & {Buildings} & {Noise} & {Trans. wires} & {Dist. wires} & {Dist. poles} & {Trans. towers} & {Fence} & {Vehicles} & {Unassigned}\\
\midrule
 Res16UNet34A & 0.793 & 69.66 & \textbf{0.99} / 97.60 & \textbf{0.99} / 98.01 & 0.62 / 45.11 & 0.80 / 66.55 & 0.99 / 97.81 & 0.86 / 75.22 & 0.63 / 45.59 & 0.86 / 74.84 & 0.83 / 70.23 & \textbf{0.88} / \textbf{78.95} & \textbf{0.28} / \textbf{16.37} \\
 Res16UNet14 & 0.797 & 70.61 & \textbf{0.99} / 97.78 & \textbf{0.99} / 98.16 & 0.62 /  44.93 & 0.82 / 69.97 & \textbf{0.99} / 97.77 & 0.86 / 74.78 & 0.68 / 51.56 & 0.94 / 88.63 & 0.81 / 68.54 & 0.83 / 71.44 & 0.23 / 13.10 \\
 Res16UNet34C & 0.843 & 76.48 & \textbf{0.99} / \textbf{98.27} & \textbf{0.99} / \textbf{98.41} & 0.88 / 78.76 & 0.83 / 70.89 & \textbf{0.99} / 98.95 & \textbf{0.92} / 84.50 & 0.74 / 59.06 & 0.95 / 88.36 & \textbf{0.85} / \textbf{73.62} & \textbf{0.85} / \textbf{73.87} & \textbf{0.28} / \textbf{16.51}\\
 Res16UNet14C & \textbf{0.845} & \textbf{77.29} & \textbf{0.99} / 98.18 & \textbf{0.99} / 98.29 & \textbf{0.92} / \textbf{84.51} & \textbf{0.84} / \textbf{72.33} & \textbf{0.99} / \textbf{99.05} & \textbf{0.93} / \textbf{86.34} & \textbf{0.75} / \textbf{60.23} & \textbf{0.96} / \textbf{91.34} & \textbf{0.85} / 73.54 & 0.84 / 72.79 & 0.24 / 13.54 \\

\bottomrule
\end{tabular}
}
\caption{Network architectures. We explore different architectures of the Minkowski Engine and find that the \textit{Res16UNet14C} backbone leads to the best semantic segmentation results in terms of macro F1 (IoU) and per-class metrics.}\label{stab:ablation-architectures}
\end{subtable}
\vspace{0.4mm}

\caption{\textbf{Ablation studies.} We provide ablation studies of the proposed dataset for different sets of point features in~\cref{stab:ablation-features}, various loss functions to handle the class imbalance (\cf~\cref{stab:ablation-losses}); different point-wise annotation strategies in~\cref{stab:ablation-annotations}; 4 different network architectures in~\cref{stab:ablation-architectures} and report the F1 / IoU metrics. The best results for each category of experiments are marked in \textbf{bold}.}\label{tab:ablations}
\end{table*}

%% file: sec/5_conclusion.tex
\vspace{-1mm}
\section{Conclusion}\label{sec:conclusion}
We present ECLAIR, a high-fidelity aerial LiDAR dataset and demonstrate how it can be used as a challenging benchmark for 3D semantic segmentation. The high-quality ground-truth labels along with pseudo labels allow benchmarking of existing point cloud semantic segmentation approaches at scale. Additionally, long-tail annotations of point clouds facilitate fine-grained semantic understanding while accommodating the uncertainty of labels. We hope that the ECLAIR dataset will introduce new challenges and stimulate the development of innovative point cloud semantic segmentation approaches that better generalize to real-world scenarios. We also aim to expand ECLAIR's capabilities to include instance segmentation annotations as future work. Furthermore, we would like to expand the ECLAIR dataset to cover larger areas. %

\boldparagraph{Acknowledgments} The authors would like to express sincere gratitude to Glenn Colvin for authorizing this research within Sharper Shape. We extend our thanks to Jaro Uljanovs for his intellectual contributions and the valuable code used in this work.
We also appreciate the technical insights provided by Khurram Gulzar, Joonas Heikkilä, Rami Piiroinen, and Jussi Sainio.
Finally, we recognize Polina Novikova for managing the data curation team, and the team members themselves – Jere Isokääntä, Kia Liljegren-Forss, Svetlana Kuznetcova, and Walter Dewald – for their essential work. Their careful review of the data significantly enhanced the quality of this research.